\documentclass[runningheads]{llncs}
\usepackage[T1]{fontenc}
\usepackage{graphicx}
\usepackage{amsmath}
\usepackage{custom_math_utils}

\begin{document}

\title{RRM: Relightable assets using Radiance guided Material extraction}
%
%
\author{Diego Gomez\inst{1}\orcidID{0009-0005-2847-8617} \and
Julien Philip\inst{2}\orcidID{0000-0003-3125-1614} \and
Adrien Kaiser\inst{2}\orcidID{0000-0002-5998-3932} \and
Élie Michel \inst{2}\orcidID{0000-0002-2147-3427}
}
%
\authorrunning{D. Gomez et al.}
%
\institute{École polytechnique \and
Adobe Research
}
\maketitle              

\DeclareGraphicsRule{.ai}{pdf}{.ai}{}

\begin{abstract}

Synthesizing NeRFs under arbitrary lighting has become a seminal problem in the last few years. Recent efforts tackle the problem via the extraction of physically-based parameters that can then be rendered under arbitrary lighting, but they are limited in the range of scenes they can handle, usually mishandling glossy scenes. We propose RRM, a method that can extract the materials, geometry, and environment lighting of a scene even in the presence of highly reflective objects. 
Our method consists of a physically-aware radiance field representation that informs physically-based parameters, and an expressive environment light structure based on a Laplacian Pyramid. We demonstrate that our contributions outperform the state-of-the-art on parameter retrieval tasks, leading to high-fidelity relighting and novel view synthesis on surfacic scenes.

\end{abstract}

\section{Introduction}

The use of fully optimizable models as 3D scene representations to address novel view synthesis problems has led in the last years to impressive results: trained on a set of multiple photographs of a scene, these models can infer unseen view angles while using as sole prior the three-dimensionality of the underlying scene. Initially based on neural network overfitting (Neural Radiance Fields~\cite{mildenhall2020nerf}), later approaches focused on improving the positional encoding fed as input to the networks (Fourier features~\cite{tancik2020fourier}, hash-grids of InstantNGP~\cite{mueller2022instant}), leading lately to neuron-free representations like TensoRF~\cite{chen2022tensorf} (when used with Spherical Harmonics decoding) or 3D Gaussian Splatting~\cite{kerbl2023gaussians}.

\begin{figure}
    \centering
    \includegraphics[width=\columnwidth]{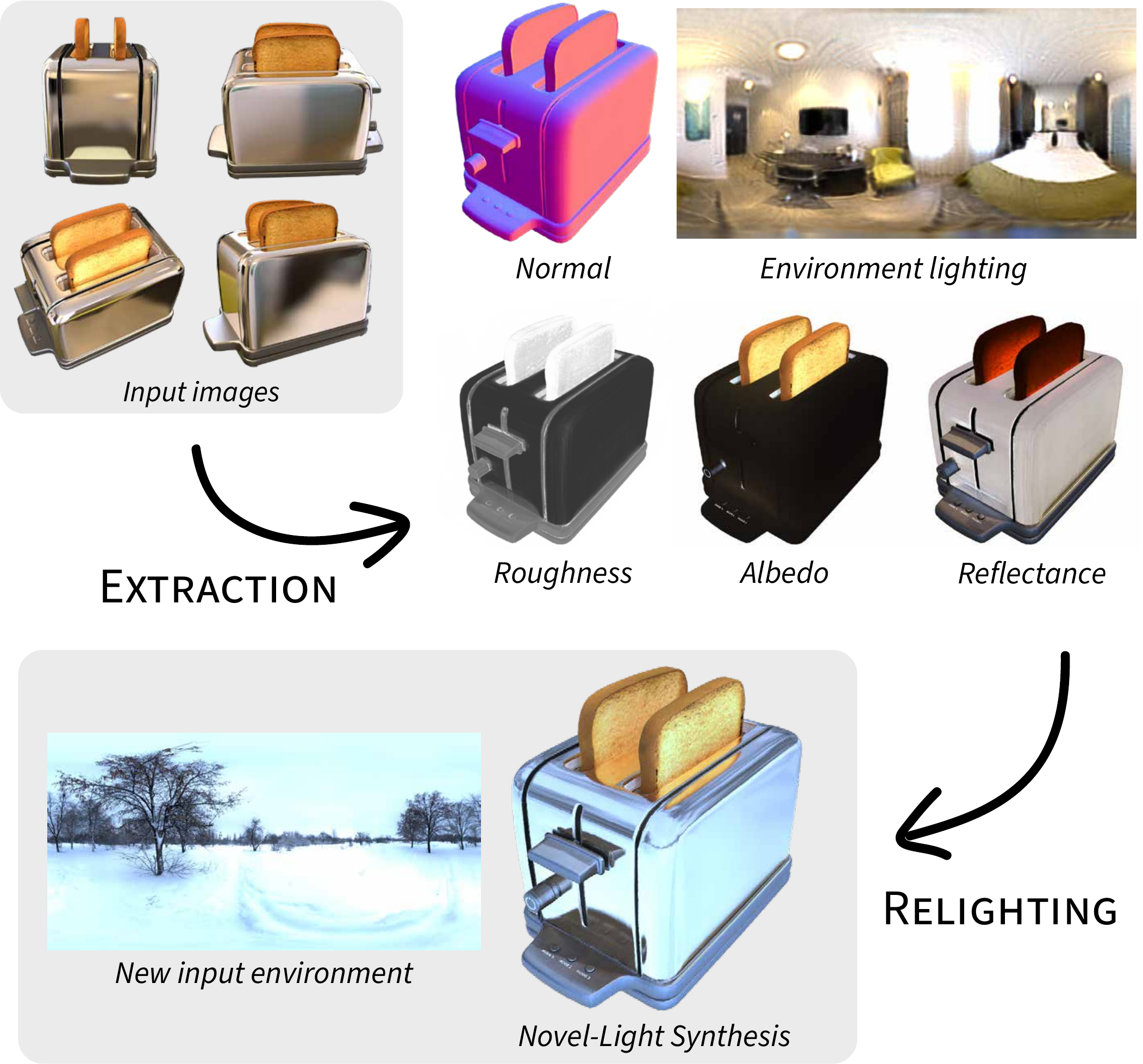}
    \caption{We take as input a collection of photographs from a scene, and extract a model with physically-based parameters from which we can set a new lighting condition. In comparison to NMF~\cite{mai2023neural} and TensoIR~\cite{jin2023tensoir}, our method is more robust to glossy materials and better handles self-reflection, as it is able to reconstruct more accurate surface normals.}
    \label{fig:teaser}
\end{figure}

The effectiveness of such overfit representation at retrieving 3D information even in the presence of transparent elements or strong specular effects makes it a strong competitor of traditional photogrammetry when it comes to acquiring 3D scenes from pictures. 
However, overfit representations usually encode only the radiance emitted by a scene, in a way that is hard to disentangle from their environment lighting at the time of acquisition. Hence a series of recent work focuses on relighting such overfit scenes, either by directly processing radiance data~\cite{toschi2023relight,xu2023renerf} or by extracting parameters compatible with physically-based 3D rendering pipelines~\cite{zhang2021nerfactor,jin2023tensoir,mai2023neural}. Our work builds on the latter, improving the extraction capability of the model thanks to a more powerful representation of environment lighting. In particular, we better reconstruct the local \textbf{surface normal} from its appearance, even in the presence of highly glossy materials.
Overall our key contributions are:

\begin{itemize}
    \item The introduction of a physically aware radiance module that extracts coarse normals and a notion of roughness, while splitting the predicted  radiance signal into view dependent and independent components.
    \item A novel way to represent environment maps based on a \textbf{Laplacian Pyramid} powered by a multiple importance sampling (MIS) algorithm that enables the retrieval of highly specular effects on complex geometry;
    \item A novel use of a radiance field as a guide to learning physically-based parameters. More specifically, a \textbf{supervision loss} on diffuse and glossy effects and the sharing of both explicit (normal, roughness) and underlying  (appearance, 3D scalar fields) parameters allows disambiguating the incoming information, leading to the extraction of high-quality parameters.
\end{itemize}
\section{Previous Work}



\paragraph{Neural Scene Representation.} Following their wide success in machine learning tasks, neural networks started being used as a means to encode high-dimensional data through overfitting. Typically applied to spatial fields (2D images, 3D signed distance fields, etc.), this approach has shown to be very good at compression and interpolation while providing fully differentiable random access look-up, hence being compatible with optimization tasks \cite{sitzmann2020implicit,martel2021acorn}. It was thus a good fit to encode 5D radiance fields, whose storage had been a longstanding challenge of computer graphics \cite{levoy96light,gortler96lumigraph}. NeRF~\cite{mildenhall2020nerf} demonstrated the use of neural network overfitting to optimize a volumetric representation whose (differentiable) render match predefined views and was soon followed by many similar approaches, progressively shifting the model's architecture towards positional encoding \cite{mueller2022instant,chen2022tensorf}.


A prominent challenge however lies in the user editing of such models. Recent efforts tackle the geometric transformations of neural representations or appearance editing of some kind \cite{verbin2022ref,NeRFshop23,yuan2022nerf}. A particularly important problem that has received attention in the past years is that of relighting \cite{zhang2021nerfactor,srinivasan2021nerv}. Some approaches attempt to tackle this by assuming known lights and parameterizing this information as an input of the model \cite{toschi2023relight,xu2023renerf}. Our work lies in the family of light-agnostic approaches, that involve leveraging inverse rendering to retrieve relightable assets \cite{jin2023tensoir,mai2023neural}.

\paragraph{Neural Material Prediction.}

In the context of material generation, there exists precedent of predicting albedo, specular, normal and roughness parameters in order to render them into the desired result, for example GAN-based methods~\cite{zhou2021adversarial}. This is related to our physically-based module (see Fig.~\ref{fig:overview}). Our physically-aware radiance module is then used to complement and inform the former. The radiance module inputs appearance related features and normals, to produce a radiance signal that we split into its view dependent and independent components.

\paragraph{Inverse Rendering.} Inverse Rendering, the translation of observed images into global geometric, material, and lighting properties is a long-standing problem in computer vision and graphics. Being an extremely under-constrained problem it requires the introduction of several priors to achieve interesting results. These priors are typically provided by the structure of the differentiable renderer used to approach the task and the underlying scene representation. Some differentiable renderers are based on rasterization \cite{hasselgren2021appearance}, point splatting \cite{yifan2019dss}, path tracing of globally illuminated meshes \cite{azinovic2019inverse}, ray marching through emissive volumes \cite{mildenhall2020nerf}. The usage of the differentiable renderer of choice dictates the priors that will be introduced to the system. In the case of NeRFs, the only prior is the 3-dimensionality of the scene, while in the case of rasterization one assumes a surfacic mesh. In our work two differentiable renderers are leveraged to perform the inverse rendering task. A physically-based one and a radiance-based one.

NeRFactor~\cite{zhang2021nerfactor} and NeRV~\cite{srinivasan2021nerv} present approaches that distill the information learned by a NeRF into a set of separate MLPs that predict geometric, visibility, and material information. These methods, however, have their limitations. NeRV requires the use of a dataset with known lighting conditions to incorporate indirect lighting information during training, whereas NeRFactor does not account for indirect lighting and self-reflections at all. TensoIR~\cite{jin2023tensoir} greatly outperforms these previous methods in the task of inverse rendering. This is achieved by leveraging TensoRF \cite{chen2022tensorf} to replace the inaccurate prediction of the visibility parameter done by its predecessors. These methods, however, are not able to tackle scenes with specular objects.

PhySG \cite{zhang2021physg} presents an inverse rendering pipeline that specializes in such objects. Nevertheless, this work does not take into account indirect illumination and thus fails at accounting for inter-reflection. The paper also restricts itself to using constant and monochrome specular BRDFs. Our method on the contrary handles both diffuse and glossy scenes, while enabling the modeling of inter-reflection and a BSDF that is spatially-varying on all components. This enables the retrieval of complex objects, such as the toaster in Fig.~\ref{fig:teaser}. 

\textbf{Neural Microfacet Fields} (NMF) for Inverse Rendering \cite{mai2023neural} takes a similar approach to extend previous works to these challenging scenes. They do this by embedding in the 3D representation introduced by TensoRF~\cite{chen2022tensorf} a microfacet representation. We however retrieve \textbf{higher quality parameters} thanks to our contributions which allow to better disentangle the incoming signal. The superior quality of the parameters we retrieve can be seen in the comparison we do in glossy scenes on the relighting task and our normal comparison quantitative results.

The recent work of NeRO \cite{liu2023nero} is able to faithfully reconstruct the geometry and the BRDF of real-life reflective objects. Their approach consists of \textbf{two stages}. First the geometry of the scene is retrieved with a neural SDF; then, with the geometry fixed, an accurate BRDF of the object is computed. Our work in contrast consists of a pipeline that is \textbf{end-to-end optimizable}. Moreover, NeRO leverages radiance fields exclusively to learn the geometry of the scene. We show that these models are capable of providing much more than reliable geometry, indeed with the proper parameterization they can provide insightful information about physical properties.






\section{Overview}

\begin{figure*}
    \centering
    \includegraphics[width=\textwidth]{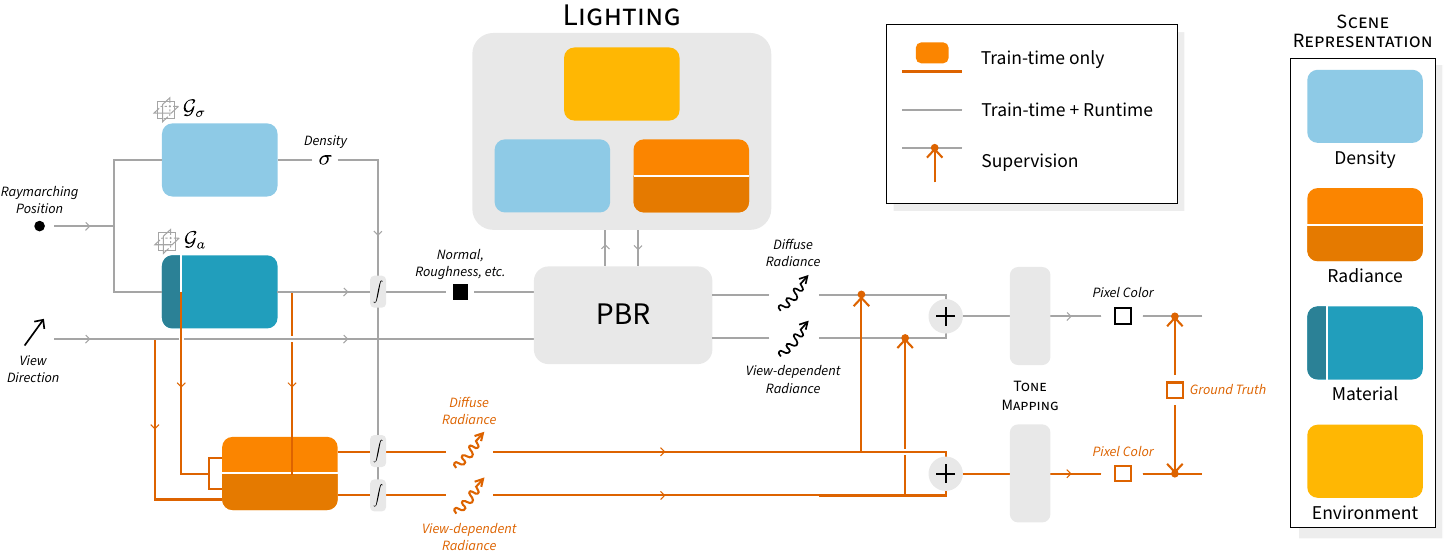}
    \caption{Overview of our model. At each ray-marching step, we evaluate a density $\sigma$, which weights physically-based material properties and radiance information as we integrate these quantities along the marched ray. Physically-based properties are processed by our PBR fixed module to compute the final radiance. Grey boxes are fixed functions, while colored boxes are the learnable scene representation. Orange boxes and arrows are used for supervision only and dropped when evaluating with a new environment lighting. See other figures for zooms of each component.}
    \label{fig:overview}
\end{figure*}

We introduce a novel method to tackle the inverse rendering problem by leveraging efficient ray marching, neural radiance fields, as well as classical light transport knowledge. This combination allows our method to retrieve high-quality geometry and material in scenes with both highly glossy and rough surfaces. Our method takes as input a set of images of the same scene, with known camera positions under one or more unknown lighting conditions, and outputs parameters that allow to render novel views using an arbitrary new environment map.

Our method is composed of multiple \textbf{learnable components} and fixed modules that we describe in Section~\ref{sec: archi} and which constitute an end-to-end trainable architecture (Fig.~\ref{fig:overview}). It includes, in particular, a physically-aware \textbf{radiance module} (Section \ref{sec: radiance}) that bootstraps the method by retrieving coarse geometry and appearance. This module then informs a \textbf{physically-based module} which learns material and fine geometry information by leveraging a physically-aware sampling algorithm (Section \ref{sec: PBR}). This sampling algorithm queries from our expressive \textbf{environment map} structure (Section \ref{sec: lighting}) based on a Laplacian Pyramid. Finally, the radiance and physically-based modules can collaborate to further disambiguate complex information in the scene (Section \ref{sec: decomposition supervision}). 

\section{Architecture} \label{sec: archi}

We represent the optimized 3D scene through 4 learnable components (see Fig.~\ref{fig:overview}.). The first 2 components are typical of NeRF-inspired methods: a \textbf{density field} encodes the coarse 3D geometry of the scene (Sec.~\ref{sec: density}), and a \textbf{radiance field} stores pre-integrated light information through the whole space (Sec.~\ref{sec: radiance}). The last 2 components are a \textbf{material field } (Sec.~\ref{sec: PBR}) and the \textbf{environment lighting} (Sec.~\ref{sec: lighting}): these encode quantities meant for physically based rendering.

The learnable components are connected together through differentiable fixed-function modules. The \textbf{Physically Based Rendering (PBR)} module turns physical material properties into radiance (Sec.~\ref{sec: PBR}), this requires the use of the \textbf{Lighting} module to estimate local irradiance (Sec.~\ref{sec: lighting}).

\subsection{Density} \label{sec: density}

The geometry of the scene is modeled as a density field. To encode this 3D scalar field, we use the TensoRF \cite{chen2022tensorf} representation. This enables highly efficient ray marching while being much faster to overfit than neuron-based models like NeRF \cite{mildenhall2020nerf}.

The TensoRF representation decomposes 3D grids into a set of vectors and matrices. For any quantity $s$, we can apply bilinear interpolation on a grid $\G_s$ to associate to any position $x \in \R^3$, we note it $s_x = \G_s(x)$.Our 3D density tensor $\mathcal{G}_\sigma$ is thus encoded using the following decomposition $\mathcal{G}_\sigma = \sum_k \sum_{m \in XYZ} v^m_{\sigma,k} \circ M^{\Tilde{m}}_{\sigma,k}$ where $v^m_{\sigma,k}, M^{\Tilde{m}}_{\sigma,k}$ is the learnable decomposition. We call $\Tilde{m}$ the corresponding complementary axes (e.g. $\Tilde{X} = Y Z$).

From this scalar field we predict the density $\sigma_{\*x}$ at a given 3D location $\*x$ as:
\begin{equation}
    \label{eq: sigma}
    \sigma_{\*x} = \mathcal{G}_\sigma(\*x)  
\end{equation}

\subsection{Physically-Inspired Radiance} \label{sec: radiance}

\begin{figure}
    \centering
    \includegraphics[width=\columnwidth]{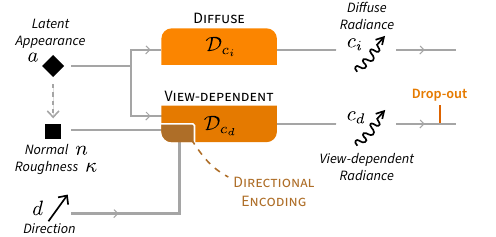}
    \caption{Our radiance component decodes the latent appearance vector coming from the material component into view-dependent and view-independent (diffuse) terms. This is done using two isolated neural networks, only one of which receives the view direction as input. The view-dependent network is made more robust to reflections by using a directional encoding based on the prediction of the material component. A drop-out on the view-dependent term ensures that the diffuse term gets as much magnitude as possible.}
    \label{fig:radiance-component}
\end{figure}

Our radiance model relies on two essential ideas (Fig.~\ref{fig:radiance-component}). The decomposition of the radiance into its view dependent and independent components. The use of the directional encoding proposed by Ref-NeRF \cite{verbin2022ref} to enable retrieval of correct geometry and density of specular objects. The latter enhances the former, not only we isolate the view dependent effects, but we model them in a manner that informs the predicted normals and roughness.

\paragraph{Latent appearance.} The radiance component inputs a latent appearance descriptor $a_x$ shared with the material component (section \ref{sec:material-component}). Similarly to TensoIR, we store this latent appearance information in a TensoRF field $
    \mathcal{G}_{\*a} = \sum_k \sum_{m \in XYZ} v^m_{\*a,k} \circ M^{\Tilde{m}}_{\*a,k} \circ b^m_{\*k}
$ The additional $b^m_{\*k}$ basis vectors express the multi-channel nature of appearance (RGB). We call $\*a_{\*x} = \mathcal{G}_{\*a}(\*x)$ the latent appearance at $\*x$.

\begin{figure}
    \centering
    \includegraphics[width=\columnwidth]{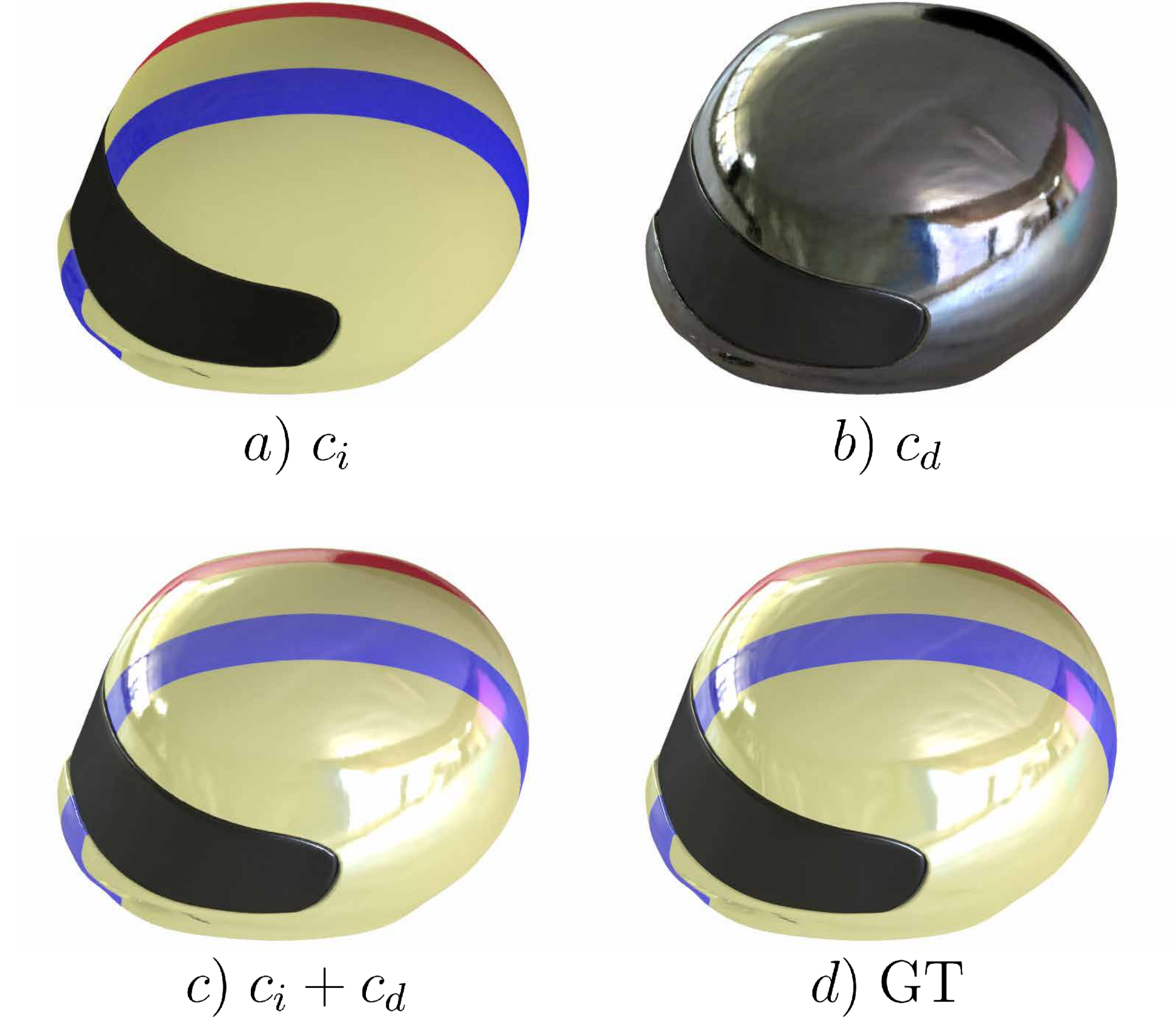}
    \caption{Visualization of our radiance decomposition as described in Fig.~\ref{fig:radiance-component} after overfitting on the \textbf{helmet} scene. This qualitatively corresponds to the diffuse and specular terms of a PBR BSDF model.}
    \label{fig:radiance-decomposition-vis}
\end{figure}

\paragraph{Radiance Decomposition.} We introduce a decomposition that allows us to isolate the view dependent and independent visual features of a scene and thus to supervise separately the diffuse and specular terms of the PBR module (Section \ref{sec: decomposition supervision}).
As illustrated in Fig.~\ref{fig:radiance-component}, the decomposition is enforced structurally by using a different decoding network for the view-independent radiance $c_i$ and the view-dependent radiance $c_d$:
\begin{equation}
\begin{aligned}
\label{eq: decomposed radiance}
    c_i(\*x) &= \mathcal{D}_{c_i}(\*a_{\*x}) \\
    c_d(\*x,\*d) &= \mathcal{D}_{c_d}(\*a_{\*x},\*d)
\end{aligned}
\end{equation}
where $\*x,\*d$ are coordinates and viewing direction of the current sample. Note that in general, we denote $\D_s$ a dense neural network and $\D_s(y)$ its prediction for a given input vector $y$. Unless otherwise stated $\D_s$ is a 3 layer MLP with ReLU activations. The input dimension is the sum of the different inputs and their respective Fourier features dimensions. The hidden dimensions are 128 for all layers. The output activation is a Softplus function with parameter $\beta_{\text{soft plus}}=3$, the output dimension is the dimension of the concatenation of the predicted quantities. This split radiance can be seen in Fig.~\ref{fig:radiance-decomposition-vis}. We then obtain the final radiance at training time with,
\begin{equation}
\label{eq: final radiance}
    c(\*x,\*d) = c_{i}(\*x) + c_{d}(\*x,\*d)
\end{equation}
Importantly, the neural network $\D_{c_d}$ contains a dropout layer during training. This is crucial for the decomposition to work (see Supp.). Contrarily to previous material prediction methods that infer albedo and specular parameters, we emphasize that we are not predicting any material properties here. Instead, through this structural choice we make the hypothesis that the radiance signal can be decomposed into view dependent and independent components. We will explore the consequences of this choice in the following.

\paragraph{Directional Encoding.} As highlighted by Ref-NeRF~\cite{verbin2022ref}, feeding the raw view direction $\*d$ to the radiance decoding network $\D_{c_d}$ leads to poor learning on glossy surfaces. We use their re-parameterization, namely to work instead with the reflected vector $\omega_r$ with respect to the predicted normal $\*n$.
In addition, we use their so-called Integrated Directional Encoding (IDE) to account for the aperture of the cone of reflection depending on an estimated roughness. We thus rewrite the view-dependent radiance in equation \ref{eq: decomposed radiance} as,
\begin{equation}
    \label{eq: ref nerf re-parameterization}
    c_d(\*x,\*d) = \D_{c_{d}}\Big( \*a_{\*x},\omega_r,\dotprod{\omega_r}{\*n_{\*x}},\textbf{IDE}(\omega_r,\kappa_{\*x}) \Big)
\end{equation}
where the normal $\*n$ and the roughness coefficient $\kappa$ are local properties predicted by our material component (Section \ref{sec:material-component}). We refer the reader to the original Ref-NeRF paper~\cite{verbin2022ref} for details about the $\textbf{IDE}$ function.

\subsection{Material}
\label{sec:material-component} 

\begin{figure}
    \centering
    \includegraphics[width=\columnwidth]{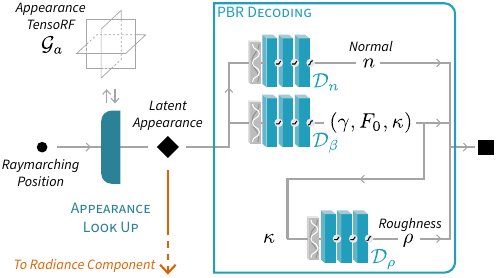}
    \caption{The material component of our scene representation consists of a look-up in a TensoRF $\G_a$ followed by a simple neural network to decode the sampled latent appearance vector into physically-based material properties. This two-stage approach enables the radiance-based component to guide the definition of a latent appearance without learning a full mapping from physically-based parameters.}
    \label{fig:material-component}
\end{figure}

We encode the physically-based material parameters in a 3D field agnostic to the current lighting condition (Fig.~\ref{fig:material-component}). These are used to \textbf{characterize the BSDF model} used in the PBR module (Section \ref{sec: PBR}): a surface normal $\normal$, an albedo $\albedo$, a reflectance (specular color) $\reflectance$ and a roughness $\roughnesspb$ parameter; and to \textbf{feed the radiance module}:  $\roughnessrad$, $\normal$.

\paragraph{A function mapping $\kappa$ to $\rho$.}

The integrated directional encoding that we import from Ref-NeRF~\cite{verbin2022ref} learns its own notion of roughness $\roughnessrad$ as a means to provide more sensibility to viewing direction in glossy areas than in rough ones. This IDE roughness is related, but however not identical to the physically-based roughness parameter $\roughnesspb$ of our BSDF model (section \ref{fig:material-component}). We could decode the physically-based roughness $\roughnesspb$ from the latent appearance independently from the IDE roughness $\roughnessrad$, but we found empirically that having two completely separate roughness parameters may lead the model to stagnate in local-maxima (Fig.~\ref{fig:kappa-to-rho-vis}). We input the Fourier features of the $\roughnessrad$ parameter and write, 
\begin{equation*}
\roughnesspb = \D_{\rho}(\roughnessrad)
\end{equation*}

The $\D_\rho$ differs from the introduced MLP in section \ref{sec: radiance}, in that the hidden dimension is 10, and the output activation is a sigmoid function, i.e. it is a much smaller network.

\paragraph{Leveraging appearance.}

We decode the remaining material quantities from the latent appearance vector $\*a_{\*x}$ previously introduced in Section \ref{sec: radiance}. More specifically:

\begin{equation} \label{eq:material}
\begin{aligned}
    \normal &= \D_{\*n}(\appfeature) \\
    (\albedo, \reflectance, \roughnessrad) &= \D_{\beta}(\appfeature)
\end{aligned}
\end{equation}

 Like radiance, material properties are evaluated at each step of the ray-marching, weighted by the local density $\sigma_{\*x}$, and integrated along the ray. When the accumulated density reaches a threshold, we feed the integrated properties to the PBR module. In other words, if the accumulated density is high enough, we compute the corresponding depth. We then transform this quantity into a surface point, for which we apply the PBR equation (described in section \ref{sec: PBR}) using the integrated parameters. In the case of a ``missed" ray, the parameters are not passed to the PBR module and thus are not updated. The corresponding pixel is given a default background color. This contributes to lowering the computational load. Moreover, similar to the radiance, this way of handling missed rays leads to a better posed learning objective; as opposed to forcing the network to predict the surrogate background value.

\subsection{Environment Lighting} \label{sec: lighting}

\begin{figure}
    \centering
    \includegraphics[width=\columnwidth]{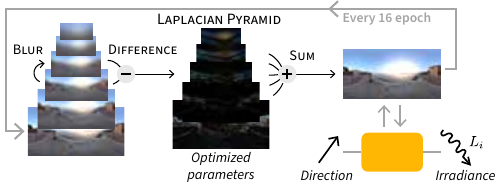}
    \caption{In our environment lighting representation, the parameters optimized by gradient descent are the levels of a Laplacian Pyramid. This multi-scale representation better learns low frequencies than raw pixels and supports high frequencies that Spherical Harmonics cannot grasp. Every 16 epochs, we re-balance the representation to ensure that it is still a Laplacian Pyramid.}
    \label{fig:envlight-component}
\end{figure}

Our fourth and last learnable component encodes the retrieved environment lighting. Similarly to concurrent work \cite{verbin2023eclipse}, our method introduces a novel way to represent environment maps. However, in our case, we rely on a Laplacian Pyramid (\textit{PoL}).

\begin{figure}
    \centering
    \includegraphics[width=\columnwidth]{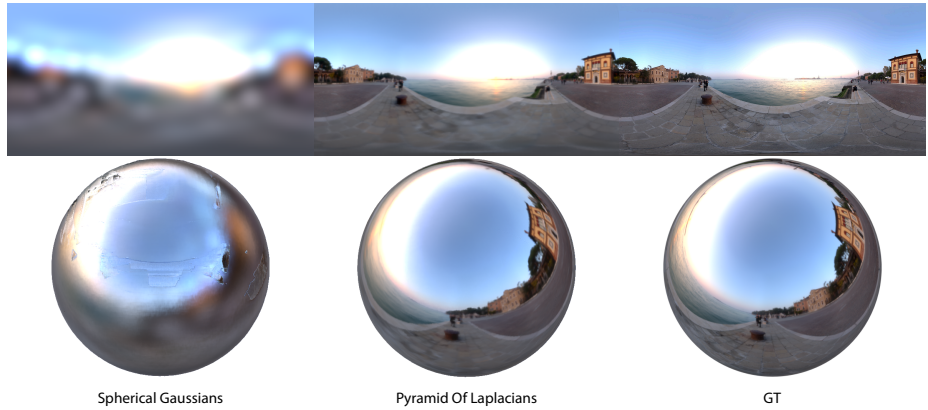}
    \caption{Reconstructed environment map and renderings for a scene made of a simple specular sphere using different representations of the environment. SG (left) fail at extracting fine details while our proposed Laplacian Pyramid (middle) gets much closer to the ground truth (right).}
    \label{fig:ablation-envmap-pol-vs-sg}
\end{figure}

\begin{figure}
    \centering
    \includegraphics[width=\columnwidth]{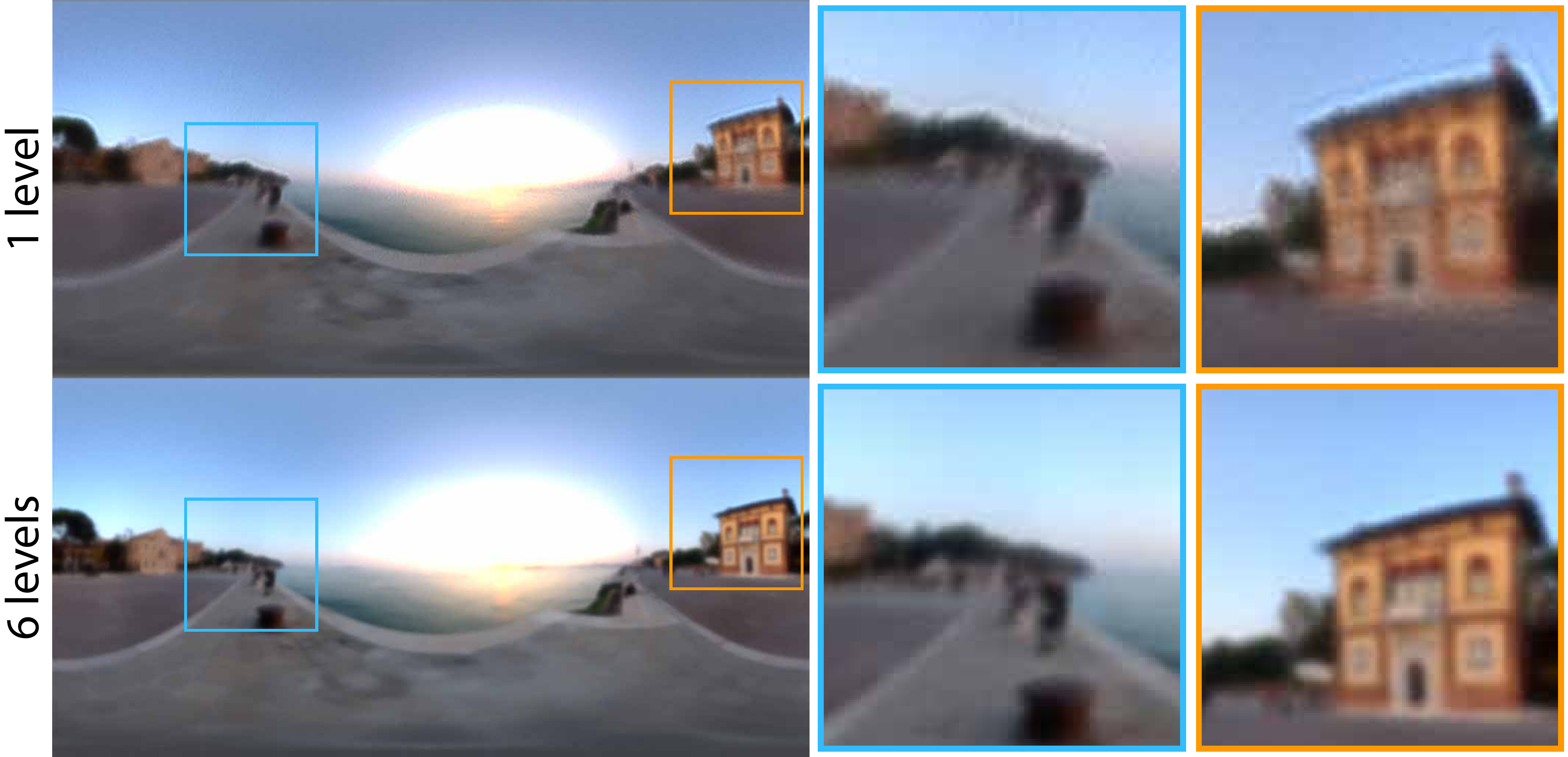}
    \caption{Directly optimizing the pixels of the envmap (top - corresponding to a single level PoL) leads to a noisier envmap with boundaries artifacts (insets). Using a 6-level PoL (bottom) provides a smoother estimate.}
    \label{fig:pol-convergence}
\end{figure}

To learn high-frequency lighting details we need a high-resolution representation, but, optimizing directly the pixels of an envmap does not allow us to leverage the correlation in nearby regions.
This leads to slow and noisy convergence of the lighting, as seen in Fig.~\ref{fig:pol-convergence}, leading to a rough optimization landscape for the other parameters such as normals.
We thus propose to optimize the levels of a Laplacian pyramid instead, allowing us to learn both low and high frequency simultaneously and to converge faster.

Given an initial envmap, we compute its Laplacian Pyramid and initialize a set of learnable parameters with the different levels.
Then during optimization, at each step, we reconstruct the envmap from the parameters and bilinearly sample the reconstruction when needed.
As parameters are optimized, there is no guarantee that the learned pyramid indeed represents the Laplacian Pyramid of the reconstructed signal.
To enforce this, at the end of every n=16 iteration, we perform a re-projection step where we reconstruct the signal from the parameters and then compute the corresponding pyramid, reassigning the value of the parameters to these levels (Fig.~\ref{fig:envlight-component}).

\begin{figure}
    \centering
    \includegraphics[width=\columnwidth]{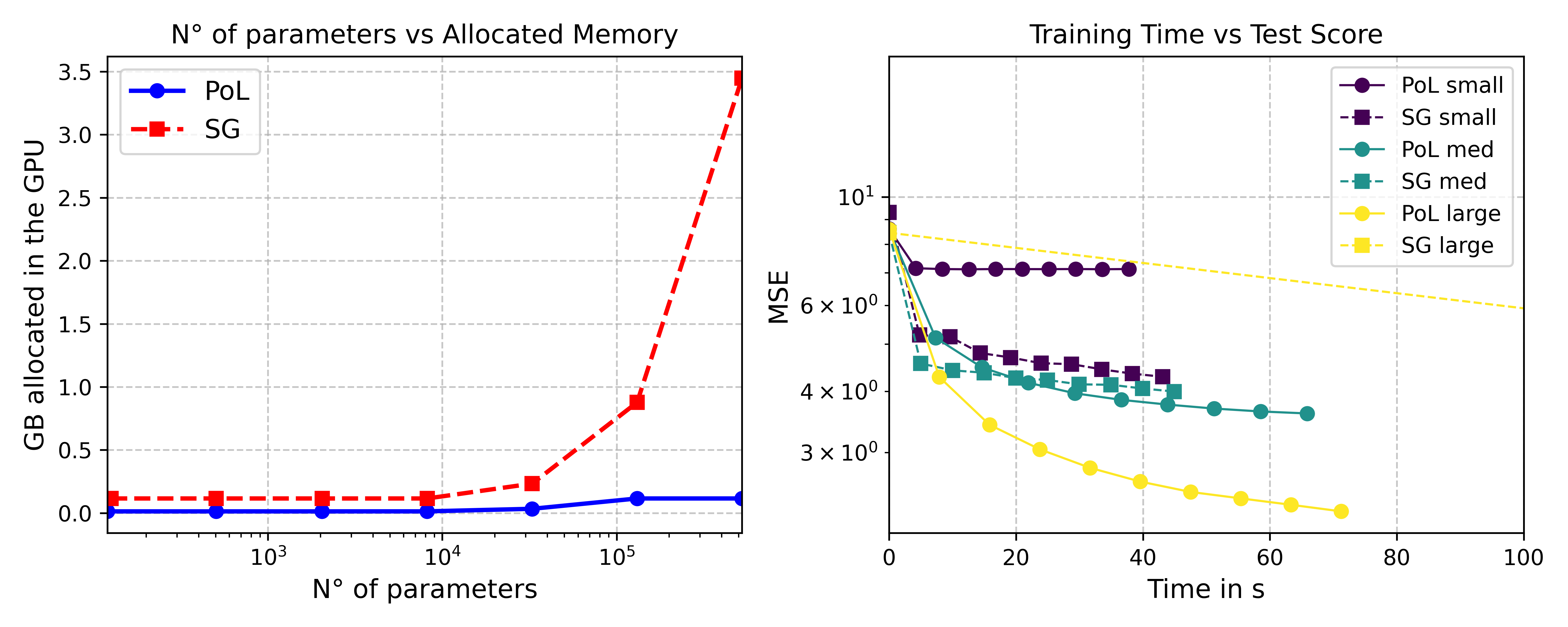}
    \caption{Experiment: Fitting the PoL and SG approaches on a given ground truth image. This simplified task gives us insight on how these methods compare. Left: we have a plot that shows the variation in allocated memory at train time as the number of parameters of each method increases. Right: Elapsed training time plotted against the test reconstruction score. Our PoL approach converges extremely fast, and larger models lead to better test results. A SG approach with a large number of lobes is challenging to train. }
    \label{fig:sg_vs_pol_image_fitting}
\end{figure}

\paragraph{PoL vs SG discussion.} Previous methods such as TensoIR~\cite{jin2023tensoir} and NeRFactor~\cite{zhang2021nerfactor} represent environment maps with Spherical Gaussians (SG). Our PoL approach is better suited than SG to learn high frequency effects efficiently. While compact, SG struggle when learning high-frequency environments. Indeed, when learning high frequency environment maps from glossy scenes one needs a large number of learnable parameters. In Fig.~\ref{fig:sg_vs_pol_image_fitting} we compare these approaches on the image overfitting task.  We can see from this figure that in such a case our PoL approach is more memory efficient and converges faster than SG.

\subsection{Physically-Based Module}
\label{sec: PBR}

\begin{figure}
    \centering
    \includegraphics[width=\columnwidth]{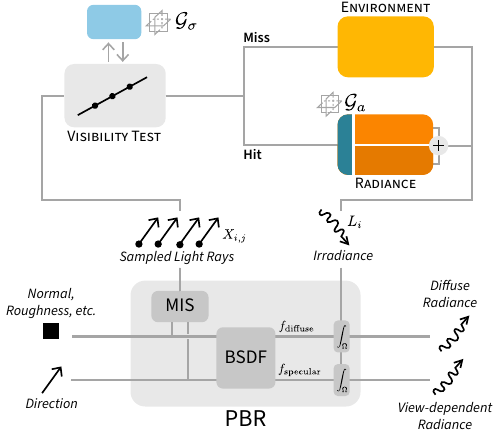}
    \caption{Our Physically-Based Rendering module uses Multiple Importance Sampling to estimate the incoming light at the shaded point. For each sampled light ray, we use either the environment or the radiance component depending on a ray marching through the density grid. For novel-light synthesis, the radiance component is replaced by a recursive call to the PBR module.}
    \label{fig:pbr-module}
\end{figure}

Concurrently to the radiance component, the PBR module also predicts a diffuse and a view-dependent radiance. However, it does it in a physically based way, that can later be user-edited, in particular by changing the lighting condition. This module is fully differentiable, so that the error gradient may flow up to the representation of the material properties and environment light. It is summarized in Fig.~\ref{fig:pbr-module}. Below we describe the fixed function renderer used. Note that, it inputs \textbf{surface parameters}, whereas our PBR module predicts \textbf{volumetric ones}. The surface point evaluated is predicted using the estimated depth, and the physically-based parameters described in equation \ref{eq:material} are aggregated along the associated marched ray to obtain the surface parameters. 

We settled for such an approach to transfer the learning flexibility of ray-marching, used by our radiance module, to the PBR module, which intends to condense light-geometry interaction to surface by fitting a surfacic BSDF model. This also mitigates the cost of indirect lighting evaluation. The accumulation-based evaluation of normals and roughness ensures graceful degradation, either when this hypothesis is wrong or while the model did not converge yet. Consistency along a ray is progressively ensured by the alignment supervision, and when the scene is indeed surfacic the density that weights accumulation is eventually null anywhere but on surfaces. 

\paragraph{Rendering.} The physically-based radiance $c_{PB}$ is computed based on the rendering equation:
\begin{equation}
    \label{eq: rendering pb}
    c_{PB}(\surfacepoint,\omega_o) = \int_{\Omega} L_i(\surfacepoint,\omega_i) f_r(\omega_o,\omega_i;\paramshat) \dotprod{\omega_i}{\normalhat}_{+} d\omega_i
\end{equation}
where $\surfacepoint$ is the surface point, $\omega_o=-\*d$ is the viewing direction, 
$L_i(\surfacepoint,\omega_i)$ is the incident illumination coming from a direction $\omega_i$, $\paramshat := (\gamma, F_0, \rho)$ are the material properties and $\normalhat$ is the normal at $\surfacepoint$. 

The BRDF $f_r$ can be split into diffuse and specular (view-dependent) terms:
\begin{equation}
    \label{eq:brdf-model}
    f_r(\omega_o,\omega_i; \paramshat)= f_{\text{diffuse}}(\albedohat) + f_{\text{specular}}(\omega_o,\omega_i;\paramshat)
\end{equation}
We integrate these terms separately as $c_{PB}^\text{dif}$ and $c_{PB}^\text{spec}$, to be able to supervise them using respectively outputs $c_i$ and $c_d$ of the radiance component. In practice, our spatially varying BRDF model is based on the Torrance–Sparrow model with a normal distribution function based on the Beckmann–Spizzichino model~\cite{beckmann1963spizzichino} (see Supp.).

\paragraph{Irradiance.} For each light ray sampled by the MIS scheme (see Supp.), we evaluate the light intensity $L_i$ coming from that direction. We leverage the efficient ray marching procedure of TensoRF \cite{chen2022tensorf} to query the incident illumination using the radiance module if the ray hits the scene, or using our environment light component otherwise.

When later evaluating the scene on a new unseen light condition for which the radiance component has no information, this radiance component is replaced by a recursive call to the PBR module, just like in a traditional ray tracer.

\section{Optimization scheme} \label{sec: optim}

Our overall optimization procedure uses a typical machine learning approach: we optimize our learnable components with a gradient descent using the common AdamW optimizer \cite{loshchilov2019decoupled} and evaluate gradients using the automatic differentiation of PyTorch. This section details some mechanisms we used to improve the convergence of our model.

\subsection{Supervision}
\label{sec: decomposition supervision}

Overall the loss we optimize is a weighted sum of the following terms: 
\begin{itemize}
    \item $l_{RF}$, $l_{PB}$ the photometric (l2) losses produced by the radiance and PB modules respectively.
    \item $l_\text{diffuse}$,$l_\text{specular}$ which we call our supervision losses on the decomposition and introduce below.
    \item $l_{\*n} = \sum_{w_j} ||n_j - n_{\sigma,j}||_2^2$,  the normal alignment loss introduced by Ref-NeRF. A loss term penalizing back-facing normals is used in addition.
    \item $l_\beta$ to ensure local smoothness loss on the different PB parameters, a Total Variation (TV) loss and $l_1$ regularization on tensor factors from \cite{jin2023tensoir,chen2022tensorf}
\end{itemize}
The radiance loss $l_{RF}$ drives the training procedure so it is the loss with the highest weight. While most of these terms were already used by TensoIR, we introduce:
\begin{align*}
    l_\text{diffuse} &= \|c_{PB}^\text{dif}(\surfacepoint) - c_i(\surfacepoint) \|^2 \text{ and, } \\
    l_\text{specular} &= \|c_{PB}^\text{spec}(\surfacepoint,\*d) - c_d(\surfacepoint,\*d) \|^2
\end{align*}
Supervising the diffuse and specular terms of the BSDF independently helps the disambiguation of intricate visual information from the input images. Although the inverse rendering problem is inherently ambiguous, this physically motivated prior results in higher quality retrieved parameters which in return improves the relighting performance.

\subsection{Radiance warm-up}

Our physically-aware radiance module is capable of learning notions of normals and roughness by itself, following Ref-NeRF. Moreover, our method heavily relies on a radiance module that has ``understood" the coarse geometry of the scene before starting the PBR procedure. Indeed, it is the radiance module that manages to process strong highlights and complex geometry in an efficient way. We therefore warm-up our radiance module by training it for 30k iterations ($\sim$1h) before enabling the PBR module.

To obtain optimal results we let our PBR module then run for another 70k iterations ($\sim$20h) on an NVIDIA Tesla T4 GPU. We have not sought to optimize this parameter and runtime in this paper. We empirically note that the time and number of iterations needed to achieve the best result in each scene significantly varies ($\sim$1h-20h).

\begin{figure}
    \centering
    \includegraphics[width=\columnwidth]{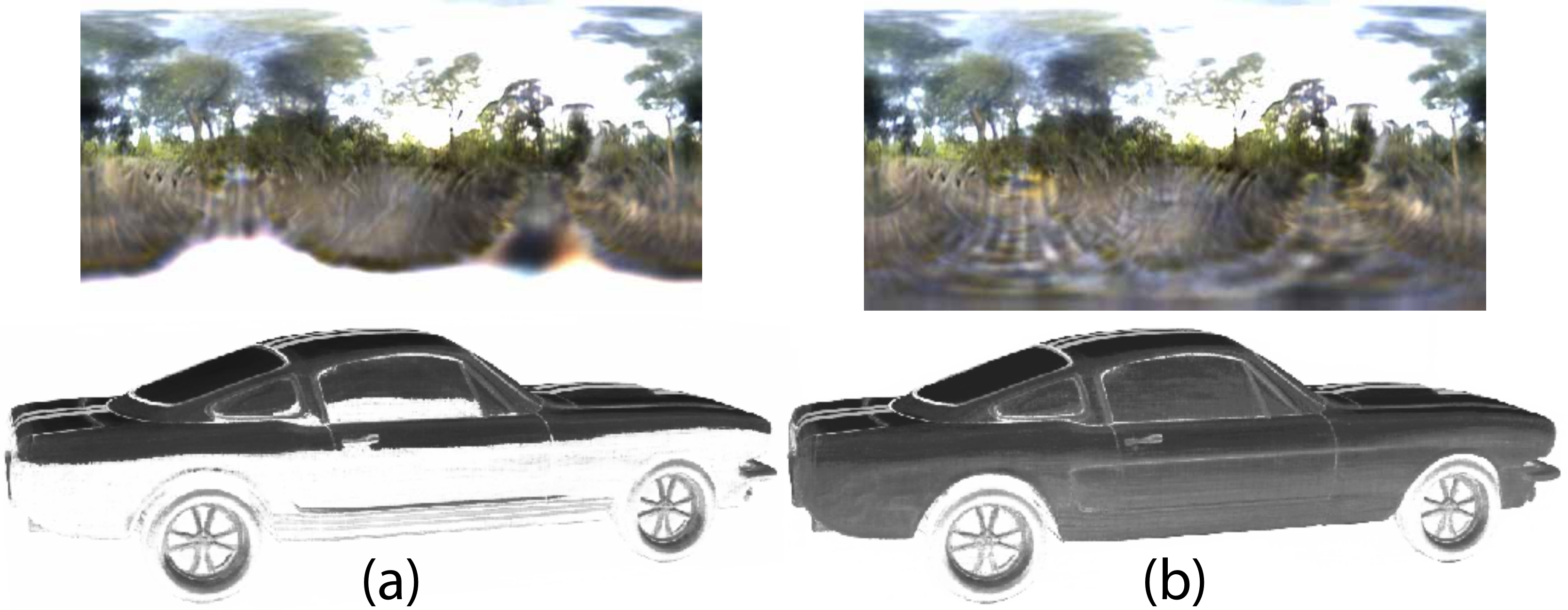}
    \caption{Visualizing $\roughnesspbhat$ with (a) separate roughness parameters, and (b) using our $\kappa \mapsto \rho$ map. The learnt environment map by each model is also visualized. We see that bad learning of the (a) model leads to loss of information in the environment map.}
    \label{fig:kappa-to-rho-vis}
\end{figure}

\section{Experiments}
\label{sec:experiments}

\begin{figure*}
    \centering
    \includegraphics[width=\textwidth]{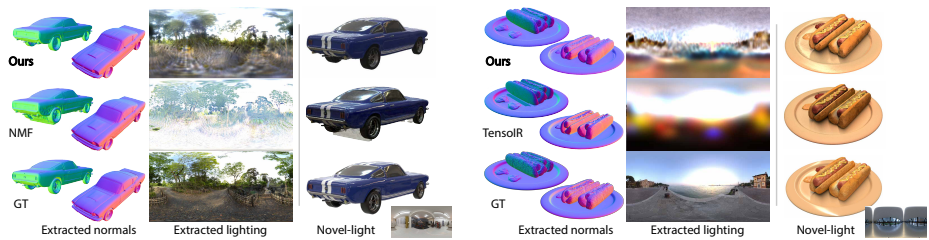}
    \caption{Comparison with TensoIR~\cite{jin2023tensoir} and NMF~\cite{mai2023neural} showing that our method better retrieves PBR parameters like normals and environment lighting, and thus leads to better relighting, especially on scenes featuring glossy surfaces.}
    \label{fig:vs-relighting}
\end{figure*}

Our method is capable of processing scenes comprised of both glossy and diffuse elements. We thus compare it to two state-of-the-art \textbf{similar} inverse rendering papers, TensoIR \cite{jin2023tensoir} which performs best on \textbf{diffuse} scenes and NMF \cite{mai2023neural} for \textbf{glossy} ones. We use  Fig.~\ref{fig:vs-relighting} to highlight the ability of our method to tackle both types of scenes. We choose in this qualitative comparison objects for which our method outperforms the aforementioned papers.

Additionally, we perform ablation studies that provide a quantitative justification for our main contributions. Our method is tested on novel-view synthesis (NVS) and relighting tasks on two synthetic datasets: the TensoIR Synthetic dataset from \cite{jin2023tensoir}, and the Shiny Blender dataset from Ref-NeRF \cite{verbin2022ref}. For any other shapes we use the pipeline introduced by NeRFactor \cite{zhang2021nerfactor}. Since these are synthetic datasets the camera information is directly extracted from blender. To evaluate performance on these two tasks we employ the standard metrics PSNR, SSIM \cite{wang2004image}, and LPIPS \cite{zhang2018unreasonable}. The \textbf{quality of our reconstructed normal} is one of the main features of our method, we measure it with the Mean Angular Error (MAE$^\circ$). 

\begin{figure}
    \centering
    \includegraphics[width=\columnwidth]{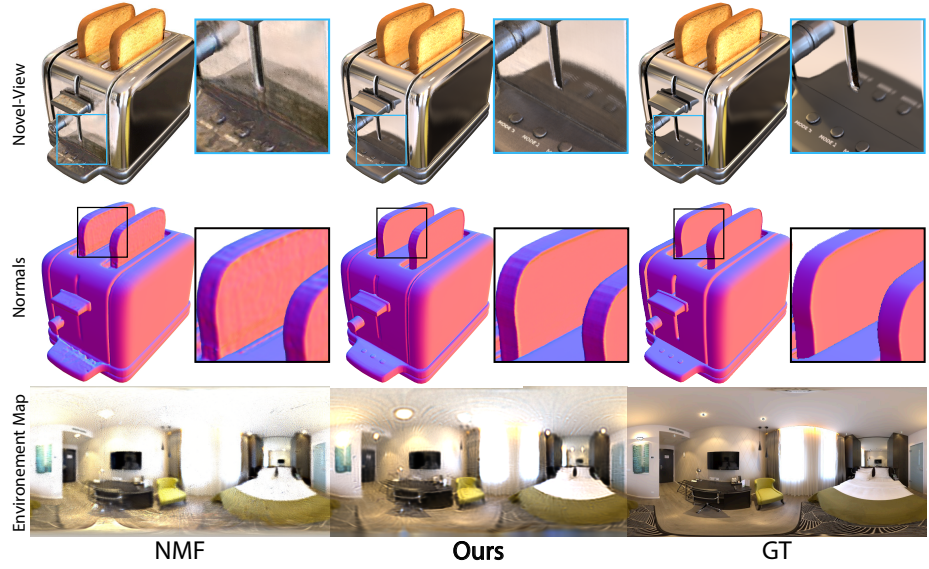}
    \caption{Comparison with NMF \cite{mai2023neural} showing that our method outperforms the state of the art in novel-view synthesis for glossy surfaces. In particular, we avoid the ghosting artifact seen in the self-reflections of NMF.}
    \label{fig:ours-vs-nmf-nvs}
\end{figure}

\paragraph{Comparison with NMF.} Our method is able to retrieve normals of much better quality than NMF, as illustrated in Fig.~\ref{fig:ours-vs-nmf-nvs} and confirmed numerically with the MAE error in Table~\ref{tab:vs-novel-view}. This table also reports that our model does not match the quantitative similarity scores performance presented by NMF on NVS task. Our model under performing on the NVS task is likely linked to the use of a neural components in NMF's BSDF model. This is not a component that is pre-trained on different materials, rather trained from scratch for each scene. This helps the overfitting of the scene, which is helpful in reconstruction. However, it does not generalize well to other light environments.

Nevertheless one can appreciate in Fig.~\ref{fig:ours-vs-nmf-nvs} that our novel views feature more consistent reflections. Indeed, we remark this qualitative improvement in multiple scenes, and the benefits of our good normal extraction becomes clear when it comes to relighting tasks: Table~\ref{tab:vs-relighting-nmf} shows that we outperform NMF on the relighting task quantitatively on the \textit{shiny blender} dataset that the NMF paper focuses on. Figure~\ref{fig:vs-relighting} shows visual examples of such relighting.

When comparing results and figures one must note that NMF uses HDR input files to train and test their model. We rather use LDR images as they are a more commonly available in practical scenarios. Moreover, we would like to highlight that the qualitative examples for NMF were directly provided by the authors of the paper. This is because, at the time of our experiments, we were unable to retrieve high quality results with the publicly available code.

\begin{figure}
    \centering
    \includegraphics[width=\columnwidth]{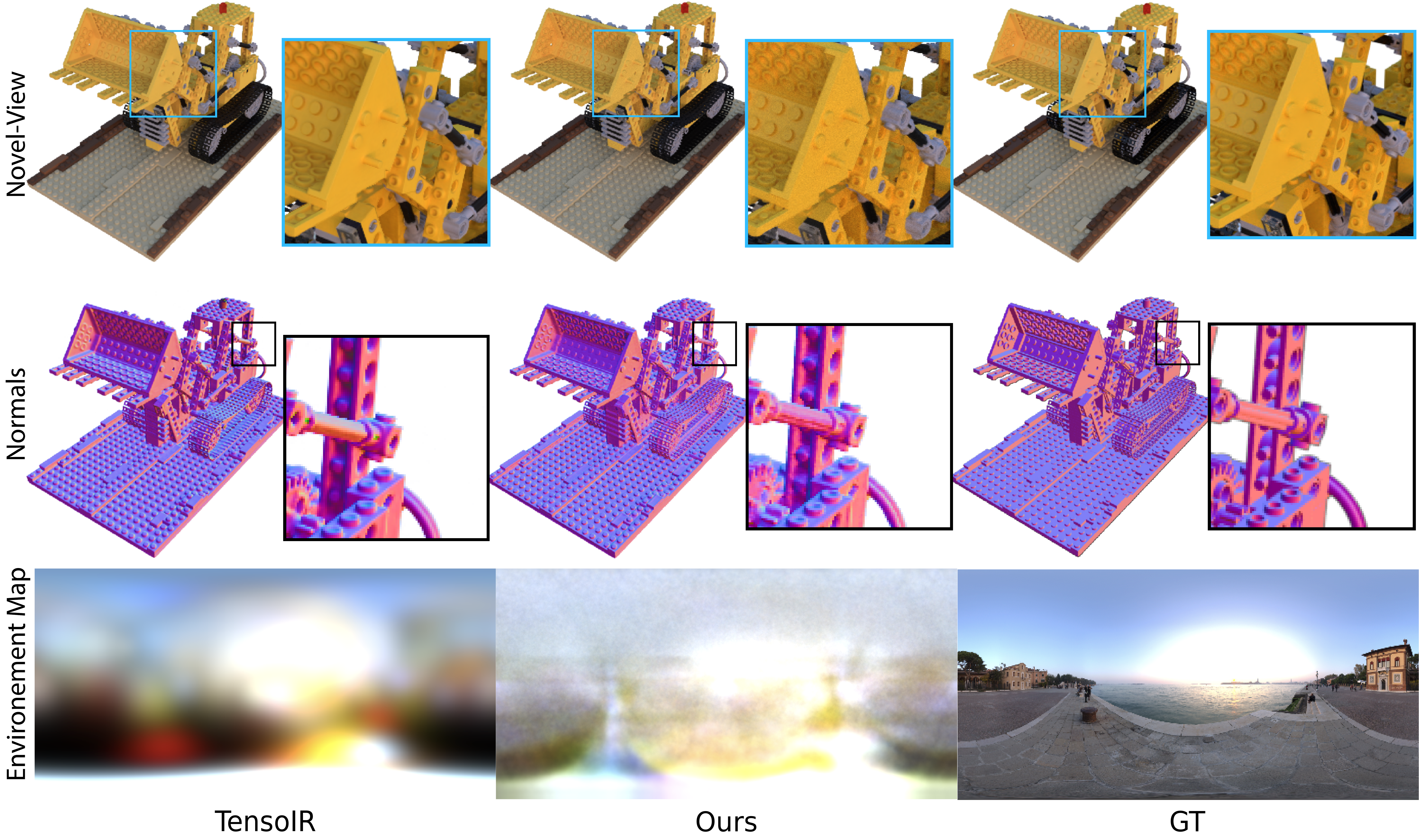}
    \caption{Comparison with TensoIR \cite{jin2023tensoir} showing that our method matches the state of the art in novel-view synthesis for diffuse surfaces. Our rendering is slightly noisier due to our importance sampling approach.}
    \label{fig:ours-vs-tensoir-nvs}
\end{figure}

\paragraph{Comparison with TensoIR.} We ran TensoIR~\cite{jin2023tensoir} on the same \textit{shiny blender} dataset as the NMF comparisons, and as we see on the bottom row of Table~\ref{tab:vs-novel-view} our method outperforms TensoIR in all metrics on NVS and normal extraction tasks. As a matter of fact, TensoIR does not perform as well on shiny scenes in general. Figure~\ref{fig:ours-vs-tensoir-nvs} shows that even on the more diffuse scenes that TensoIR targets, we \textbf{maintain} PBR quality while slightly improving on the retrieved normals. In this figure we can see however, that the MIS algorithm comes with some undesired noise. Indeed, whereas using fixed light sampling limits the rendering of shiny materials, it produces less noisy results for diffuse objects than MIS for the same number of samples.

TensoIR uses the learned radiance as a proxy for indirect lighting during training. This however, cannot be done during testing on different lighting conditions. Thus, the authors decided to omit indirect lighting during their evaluation process. Our framework allows for evaluation of indirect lighting, thanks to a \textbf{slightly} more sophisticated renderer. This is what we use for our teaser (figure \ref{fig:teaser}), which explains why we can see reflections on the relit scene. For a proper comparison on the relighting task we restrict our comparison to the TensoIRSynthetic dataset and do not compute indirect lighting. Table \ref{tab:vs-relighting-tensoir} reports that our method is slightly over performed by TensoIR, but manages to \textbf{maintain} high quality results by achieving better results than other well-established methods such as NeRFactor~\cite{zhang2021nerfactor} and InvRender~\cite{zhang2022modeling}. 
Given that important parameters, such as the normals, are better predicted by our method we can attribute the slight difference in performance to our \textbf{noisier} rendering pipeline, due to our use of MIS to sample light directions as previously discussed.

\paragraph{Ablations.} Table~\ref{tab:vs-novel-view} shows results on a novel view synthesis task for different ablations of our pipeline. Each ablation removes one of the building blocks of the method. First, ``w/ separate $\rho$" learns the PBR $\roughnesspb$ directly as an output of $\D_{\beta}$ in equation \ref{eq:material}, independently from the IDE roughness $\roughnessrad$, ``w/o Decomposition" replaces our decomposition introduced via equation \ref{eq: decomposed radiance} by the same radiance map used by TensoRF. Finally, ``w/o Supervision" omits the losses introduces in section \ref{sec: decomposition supervision}, $l_\text{diffuse}$,$l_\text{specular}$.

We can see in Table~\ref{tab:vs-relighting-ablation} our ablation results for the relighting tasks. Although ``Ours separate $\rho$" allows for slightly better normal reconstruction, we see ultimately the benefit of our $\kappa \mapsto \rho$ map in relighting scores. Indeed, this mapping is important to better retrieve the roughness in ambiguous scenes. In figure \ref{fig:kappa-to-rho-vis}, we can see how utilizing our $\kappa \mapsto \rho$ mapping allows to leverage the notion of roughness learned by the radiance to avoid losing information. The ``w/o Supervision" seems to yield quite similar results to our method. However, we can see in the detailed tables presented in our supplemental section, that supervision helps in scenes where the decomposition is the cleanest (figure \ref{fig:radiance-decomposition-vis}) such as helmet or toaster. In diffuse scenes our method without supervision performs better. One could alleviate this by setting a smaller weight for the supervision. We decided to use common weights among our tests to present a fair comparison. 

Our Laplacian Pyramid model for environment lighting is compared to the mixture of SG used by TensoIR in section \ref{sec: lighting}. Although very flexible for learning rough lighting, it eventually fails at grasping the fine details of the environment. This is not a problem for rough objects, but as shown on the toy sphere example, it makes it impossible to properly learn very glossy materials and leads to artifacts in the albedo and roughness. Lastly, Fig~\ref{fig:pol-convergence} highlights that the multi-scale approach of the Laplacian Pyramid benefits to the final quality of the environment reconstruction. 

\begin{table}
    \label{tab:vs-novel-view}
    \caption{Quantitative comparison on the shiny blender test. The PSNR, SSIM and LPIPS scores measure the similarity between novel view synthesis and ground truth under the same light condition. The MAE score characterizes the reconstruction of the geometry, which is independent from any lighting. We highlight the \first{best} and the \second{second best} scores.}
    \centering
    \begin{tabular}{|l|l|l|l|l|}
    \hline
         & \textbf{PSNR} $\uparrow$ & \textbf{SSIM}$\uparrow$ & \textbf{LPIPS} $\downarrow$ & \textbf{MAE}$\downarrow$ \\ \hline
        Ours &  31.635 & 0.941 & 0.098 & 2.262 \\ \hline
        \wrap{w/ separate $\rho$} & 32.009 & 0.945 & 0.098 & \first{2.197} \\ \hline
        \wrap{w/o Decomposition} & 31.235 & 0.936 & 0.108 & 2.809 \\ \hline
        \wrap{w/o Supervision} & \second{32.289} & \second{0.947} & 0.098 & \second{2.260} \\ \hline
        \hline
        TensoIR & 31.296 & 0.939 & \second{0.089} & 4.390 \\ \hline
        NMF & \first{33.599} & \first{0.958} & \first{0.046} & 3.659 \\ \hline
    \end{tabular}
\end{table}

\begin{table}
    \label{tab:vs-relighting-nmf}
    \caption{Quantitative comparison of the similarity between relighted scenes and ground truth on the shiny blender dataset. We highlight the \first{best} and the \second{second best} scores.}
    \centering
    \begin{tabular}{|c|c|c|c|}
    \hline
          & \textbf{PSNR} $\uparrow$ & \textbf{SSIM}$\uparrow$ &\textbf{LPIPS} $\downarrow$\\ \hline
        \textbf{Ours} & \first{25.838} & \first{0.925} & \first{0.101} \\ \hline
        NMF & \second{25.502} & \second{0.916} & \second{0.113} \\ \hline
        NVDiffRec & 20.686 & 0.8312 & \third{0.191} \\ \hline
        NVDiffRecMC & \third{22.196} & \third{0.874} & 0.2158 \\ \hline
    \end{tabular}
\end{table}

\begin{table}
    \label{tab:vs-relighting-ablation}
    \caption{Quantitative comparison ablation of relighting task. We highlight the \first{best} and the \second{second best} scores.}
    \centering
    \begin{tabular}{|c|c|c|c|}
    \hline
          & \textbf{PSNR} $\uparrow$ & \textbf{SSIM}$\uparrow$ &\textbf{LPIPS} $\downarrow$ \\ \hline
        Ours & \first{25.261} &  \second{0.924} & \second{0.096} \\ \hline
        w/ separate $\rho$ & \third{25.113} & \third{0.921} &  \third{0.097} \\ \hline
        w/o Decomp & 24.886 & 0.915 & 0.105 \\ \hline
        w/o Supervision & \second{25.179} & \first{0.926} & \first{0.091} \\ \hline
    \end{tabular}
\end{table}
\begin{table}
    \label{tab:vs-relighting-tensoir}
    \caption{Quantitative comparison between relighted scenes on the TensoIR Synthetic dataset. We highlight the \first{best} and the \second{second best} scores.}
    \centering
    \begin{tabular}{|c|c|c|c|}
        \hline
          & \textbf{PSNR} $\uparrow$ & \textbf{SSIM}$\uparrow$ &\textbf{LPIPS} $\downarrow$ \\
        \hline
        Ours & \second{28.144} & \second{0.929} & \second{0.085} \\ \hline
        TensoIR & \first{28.58} & \first{0.944} & \first{0.081} \\ \hline
        NeRFactor & 23.383 & \third{0.908} & 0.131 \\ \hline
        InvRender & \third{23.973} & 0.901 &  \third{0.101} \\ \hline
    \end{tabular}
\end{table}
\section{Discussion}

We have introduced a novel and powerful approach to tackle the ambiguous problem of inverse rendering. Using a set of input images with their respective camera information we can generate the geometry, environment illumination, and material properties of the scene. We leverage both volumetric rendering through the use of our radiance module, and physically-based rendering. 

\subsection{Properties and insights}

We have shown in our quantitative and qualitative tests that our method is suited for a wide a range of effects. It can capture scenes composed of complex geometry (Fig.~\ref{fig:ours-vs-tensoir-nvs}) and it excels at tackling glossy surfaces (Fig.~\ref{fig:ours-vs-nmf-nvs}).

In contrast to state-of-the-art methods such as NeRO \cite{liu2023nero}, which advocate for processing information through separate geometry and material stages, our method tackles the problem with a single-stage, end-to-end optimizable architecture. One of NeRO's limitation according to the authors is the failure of retrieving subtle geometrical details. Even though our method is not put to the test on real-life data, the quality of the normals we achieve on our tests seem to indicate that our model does not have this limitation.

Our contributions advocate for similar methods. That is, end-to-end optimizable radiance guided approaches to inverse rendering. We have showed on this paper that NeRFs can provide good coarse features. Indeed, we achieve state-of-the-art extraction of normals. This is a crucial step in scene understanding, since given a neat normal, the learning of the other parameters is better constrained. Furthermore, NeRFs can help to disambiguate the inverse rendering problem, our radiance decomposition leads to an increase in performance in all our tests, and using the roughness predicted by the radiance can help to escape local extrema. As opposed to methods that focus on rendering predicted PB parameters (NMF \cite{mai2023neural}, NeRO\cite{liu2023nero}) we show that dual rendering approaches can help to better condition the ill posed problem of inverse rendering.

\subsection{Limitations}

Our model is not without limitations; each of its components may be limiting in some situation. If the radiance component cannot ``understand" the scene enough to initiate the extraction, the PBR module cannot help it getting out of strong local minima. Like the methods we compare to, we focused our tests on surfacic scenes, with no semi-transmissive volumes. If the scene contains light scattering effects that our PBR module cannot replicate (e.g., subsurface scattering, iridescence, etc.), it will only try to fit its BSDF model, which may notably mess up with normal extraction. Our model will not perform optimally. Moreover, strong inter-reflections are very hard to properly extract, and our method fails when the far lighting assumption is not met (See Supp.).

Even if we could perfectly reconstruct the geometry of the scene, the environment light and material properties are only retrieved up to a multiplicative ambiguous parameter: multiplying the lighting by a fixed factor can be balanced by globally reducing the albedo and reflectance. More generally, it is difficult to find common ground for comparison in the space of physically-based parameters, as they rely on different BSDF models, some of which are even partially learned~\cite{mai2023neural}. This is why we focused our efforts on normals. Ultimately, the choice of BSDF model depends on downstream use of the extracted 3D. The quantitative evaluation of the environment reconstruction is also a question: errors in areas that do not contribute to the rendered images, or only contribute to rough surfaces should not be penalized in the same way as sharp reflections.

\subsection{Future Work}

Each of our components can be individually improved. Most insights of our system are not specific to the TensoRF model for positional encoding, so other learnable representations could be used. Our environment light component could be augmented in order to support not only far directional light, but also near distance light sources, like out-of-frustum surfaces, which we would typically meet when applying our system to non-synthetic images. We could explore the behavior of our approach in presence of transmissive surfaces, leading to multiple calls to the PBR module per camera ray, or its interaction with rasterization-based inverse rendering like 3D Gaussian Splatting paper \cite{kerbl2023gaussians}. Lastly, our method still has a lot of room to be optimized. There exist optimized NeRF libraries from which our method could benefit from \cite{li2022nerfacc}.


\begin{credits}
\subsubsection{\ackname} We wish to warmly thank the authors of TensoIR~\cite{jin2023tensoir} and NMF~\cite{mai2023neural} for providing details and results beyond what was available in the original publications.

\end{credits}

%
%
%
\bibliographystyle{splncs04}
\bibliography{bib_CGI}

\end{document}


\input{utils}

\title{Supplementary Material | RRM: Relightable assets using Radiance guided Material extraction}
%
%
\author{Diego Gomez\inst{1}\orcidID{0009-0005-2847-8617} \and
Julien Philip\inst{2}\orcidID{0000-0003-3125-1614} \and
Adrien Kaiser\inst{2}\orcidID{0000-0002-5998-3932} \and
Élie Michel \inst{2}\orcidID{0000-0002-2147-3427}
}
%
\authorrunning{D. Gomez et al.}
%
\institute{École polytechnique \and
Adobe Research
}
\maketitle              

\DeclareGraphicsRule{.ai}{pdf}{.ai}{}

\section{BSDF Model}

The function $f_r$ in Section "PBR module" of the main paper is our spatially varying BRDF model based on the Torrance–Sparrow model with a normal distribution function based on the Beckmann–Spizzichino model. We write it as,

\begin{equation}
    \label{eq:brdf-model}
    f_r(\omega_o,\omega_i; \paramshat)= f_{\text{diffuse}}(\albedohat) + f_{\text{specular}}(\omega_o,\omega_i;\paramshat)
\end{equation}
Where,
\begin{align*}
    f_{\text{diffuse}}(\albedohat) &= \dfrac{\albedohat}{\pi} \\
    f_{\text{specular}} (\omega_o,\omega_i; \paramshat) &= \frac{
D(\*h,\roughnesspbhat) 
G(\omega_o,\omega_i,\roughnesspbhat)
F_{\text{Schlick},\surfacepoint}(\omega_o,\*h)}
{4 
\dotprod{\normalhat}{\omega_i}
\dotprod{\normalhat}{\omega_o}}
\end{align*}
where $\*h$ is the half-vector between $\omega_o$ and $\omega_i$, $D$ 
 the Beckmann–Spizzichino model \cite{beckmann1963spizzichino},
 $$
 G(\omega_o,\omega_i,\roughnesspbhat) = G1(\omega_o,\roughnesspbhat) G1(\omega_i,\roughnesspbhat)
 $$
 
with $G_1$ the Smith’s masking-shadowing function, and
$$
F_{\text{Schlick},\surfacepoint}(\omega_o,h) = \reflectancehat + (1-\reflectancehat)(1 - \dotprod{\omega_o}{\*h}^5)
$$
the Schlick. \cite{schlick1994inexpensive} approximation for the Fresnel term.

\paragraph*{Multiple Importance Sampling.} The BRDF sampling we implement involves the sampling of the normal distribution function 
\begin{equation*}
D_\omega(\omega_h) = \frac{D(\omega_h) G_1(\omega_h) \dotprod{\omega}{\omega_h}_+}{\cos \theta}   
\end{equation*}
To perform light sampling, we retrieve the environment(s) map(s) at the current iteration. For a given environment map we map it into a 2D probability distribution by considering for each cell the corresponding normalized light intensity (sum of the RGB channels). Each cell is associated to a light direction given our far away lighting assumption. We retrieve the pdf of a given direction in polar coordinates as,
\begin{equation*}
p_{\text{light}}(\theta,\phi) = \frac{p_{\text{envmap}}(u,v) N_l}{2 \pi^2 \sin(\theta)} 
\end{equation*}
where $(\theta,\phi)$ is the polar coordinates of the direction that we query, $(u,v)$ is the corresponding coordinates of the associated cell and $N_l$ is the total number of pixels in our envmap ($N_l=512\times1024$ in our case). 

The MIS algorithm is thouroughly explored in \cite{veach1995optimally}, let us give a brief overview. We use this method to evaluate the function 
\begin{equation*}
g(\omega) = L_i(\surfacepoint,\omega) f_r(\omega_o,\omega;\paramshat) \dotprod{\omega}{\normalhat}_{+}
\end{equation*}
Consider $n$ sampling strategies with corresponding density functions $\{p_i \}_{i=1}^n$. Call $X_{i,j}$ the $j$-th sample from strategy $i$. The \textit{multi-sample estimator} is given by,
\begin{equation}
    \label{eq:mis}
    F = \sum_{i=1}^n \frac{1}{n_i} \sum^{n_i}_{j=1} w_i(X_{i,j}) \frac{g(X_{i,j})}{p_i(X_{i,j}}
\end{equation}
With a $w_i$ the associated weighing function of strategy $i$. Note that it must follow,
\begin{itemize}
    \item $\sum_{i=1}^n w_i(x) = 1$ for any $x$ s.t. $g(x) \neq 0$.
    \item $w_i(x) = 0$ whenever $p_i(x) = 0$
\end{itemize}
Finally, we apply the above we our two sampling strategies: light and BRDF sampling.

\paragraph{Sampling Light directions.} The integral in the rendering equation can only be evaluated numerically in general. Previous radiance field relighting methods such as TensoIR~\cite{chen2022tensorf} or NeRFactor~\cite{zhang2021nerfactor} use a uniform discretization of the set $\Omega$ of incoming directions, but decades of light transport research point to the use of Monte Carlo methods to retrieve high-frequency glossy effects. We thus employ a Multiple Importance Sampling (MIS) scheme \cite{veach1995optimally}, using both a prior based on the environment light and one based on the BRDF in order to select good light rays.
We compare qualitatively our method with and without MIS in figure \ref{fig:fixed-vs-mis} and give more details about our importance sampling in the supplementary material.

\section{Radiance Decomposition details}

In order to enforce a meaningful decomposition of the radiance into view-independent and view-dependent components, $c_i$ and $c_d$, we introduce a simple yet effective strategy.
During training, we apply a dropout layer to the output of the view-dependent network $\D_{c_d}$ with a probability of $p=0.01$, meaning we drop the view-dependent component of the radiance for 1\% of the samples. The 1\% of the time where it is not present, all the gradient of this loss goes towards the diffuse component effectively pushing it to reconstruct the full signal and thus capture as much radiance as it can and helps to disambiguate the split. Without this dropout there are infinitely many solutions to $c_d+c_i=O$ with $O$ the observed values for a given 3D point.

\section{Extra-Properties}

We inherit the ability to input multiple unknown lighting conditions from TensoIR \cite{jin2023tensoir}, as shown in Fig.~\ref{fig:ablation-multi-light}. We did not explore this feature in this paper since it is not part of our contributions. It is nonetheless to be expected that the aggregation of different light conditions will lead to better results since the ambiguity in the system is greatly reduced.

\begin{figure}
    \centering
    \includegraphics[width=\columnwidth]
    {figures-src/multi_view_ablation.ai}
    \caption{Similarly to TensoIR~\cite{jin2023tensoir}, our method gradually improves as we increase the number of unknown light settings under which input images are taken.}
    \label{fig:ablation-multi-light}
\end{figure}

\begin{figure}
    \centering
    \includegraphics[width=\columnwidth]{figures-src/fixed_vs_mis.ai}
    \caption{Novel-View Synthesis and retrieved environment maps for models trained using Uniform (left) or Multiple Importance Sampling (middle). Uniform sampling limits the range of glossy effects the model can learn as seen when compared to the Ground Truth (right).}
    \label{fig:fixed-vs-mis}
\end{figure}

\begin{figure}
    \centering
    \includegraphics[width=\columnwidth]{figures-src/vis_fail.ai}
    \caption{Our model fails to retrieve the specular effects on the coffee scene, notably on the cup. This is a hard scene due to the strong near-lighting effects. We note that our model manages to ``understand" the scene better than NMF \cite{mai2023neural} which results in more plausible normals. }
    \label{fig:fail-case}
\end{figure}

\section{Quantitative results detail}

Tables \ref{tab:vs-nvs-psnr},\ref{tab:vs-nvs-ssim}, \ref{tab:vs-nvs-lpips} and \ref{tab:vs-nvs-mae} give the results of the novel view synthesis test. A notable thing is that the no supervision model seems to yield heavier assets, unfortunately we did not perform a formal study to confirm this. We see however that in the hotdog scene this ablation fails with an OOM error message.

\begin{figure}
    \centering
    \includegraphics[width=\columnwidth]{figures-src/xyz_exp.ai}
    \caption{If we do not give an encoding of the location $\*x$ our radiance method fails in some scenes with complex lighting information.}
    \label{fig:xyz-exp}
\end{figure}

\begin{table*}[t]
    \caption{PSNR $\uparrow$ on novel view synthesis tasks (higher is better).}
    \centering
    \begin{tabular}{|c| c c c c c c c|c c c c|}
        \hline
         \wrap{Method} & Teapot& Toaster & Car & Ball & Coffee & Helmet & Materials & Lego & Hotdog & Ficus & Armadillo \\
        \hline
        \wrap{Ours} & 41.75 & 26.33 & 29.77 & 36.07 & 29.71 & 30.74 & 29.04 & 31.90 & 25.22 & 29.57 & 37.89 \\
        \hline
        \wrap{Ours separate $\rho$} & 41.63 & 26.00 & 29.66 & 35.73 & 30.44 & 30.71 & 29.46 & 31.88 & 29.22 & 29.57 & 37.80 \\
        \hline
        \wrap{Ours No Decomposition} &  41.43 & 22.44 & 28.07 & 36.11 & 30.41 & 28.57 &  29.45 & 31.89 & 27.77 & 29.65 & 37.79 \\
        \hline
        \wrap{Ours No Supervision} & 41.65 &  26.26 & 29.65 & 35.93 & 29.55 & 31.16 & 29.39 & 31.95 & OOM & 29.68 & 37.67 \\
        \hline
        \hline
        \wrap{TensoIR} & 42.44 & 19.69 & 26.52 & N/A & 31.22 & 25.94 & 26.80 & 34.700 & 36.820 & 29.780 & 39.050 \\
        \hline
        \wrap{NMF} & 45.29 & 27.52 & 30.28 & 38.41 & 31.47 & 34.38 & 31.19 & 32.98 & 35.23 & 29.24 & N/A\\
        \hline
    \end{tabular}
    
    \label{tab:vs-nvs-psnr}
\end{table*}

\begin{table*}[t]
    \caption{SSIM $\uparrow$ on novel view synthesis tasks (higher is better).}
    \centering
    \begin{tabular}{|c| c c c c c c c|c c c c|}
        \hline
         \wrap{Method} & Teapot& Toaster & Car & Ball & Coffee & Helmet & Materials & Lego & Hotdog & Ficus & Armadillo \\
        \hline
        \wrap{Ours} & 0.9843 & 0.9217 & 0.9429 & 0.9786 & 0.9100 & 0.9075 & 0.9452 & 0.9591 & 0.8957 & 0.9711 & 0.9822 \\
        \hline
        \wrap{Ours separate $\rho$} & 0.9844 & 0.9173 & 0.9401 & 0.9779 & 0.9171 & 0.9070 & 0.9481 & 0.9586 & 0.9352 & 0.9699 & 0.9820 \\
        \hline
        \wrap{Ours No Decomposition} &   0.9874 & 0.8635 & 0.9211 & 0.9798 & 0.9122 & 0.8875 & 0.9486 &  0.9560 & 0.9383 & 0.9702 & 0.9817 \\
        \hline
        \wrap{Ours No Supervision} & 0.9881 &  0.9196 & 0.9428 & 0.9784 & 0.9115 & 0.9128 & 0.9471 & 0.9577 & OOM & 0.9703 & 0.9807 \\
        
        \hline
        \hline
        \wrap{TensoIR} & 0.9953 & 0.8222 & 0.9220 & N/A & 0.9656 & 0.9107 & 0.9274 & 0.968 & 0.976 & 0.973 & 0.985\\
        \hline
        \wrap{NMF} & 0.996 & 0.917 & 0.951 & 0.983  & 0.960 & 0.969 & 0.959 & 0.963 & 0.964 & 0.952 & N/A\\
        \hline
    \end{tabular}
    
    \label{tab:vs-nvs-ssim}
\end{table*}

\begin{table*}[t]
    \caption{LPIPS $\downarrow$ on novel view synthesis tasks (lower is better).}
    \centering
    \begin{tabular}{|c| c c c c c c c|c c c c|}
        \hline
         \wrap{Method} & Teapot& Toaster & Car & Ball & Coffee & Helmet & Materials & Lego & Hotdog & Ficus & Armadillo \\
        \hline
        \wrap{Ours} & 0.0449 & 0.1473 & 0.0704 & 0.1350 &  0.1672 & 0.1921 & 0.0675 & 0.0771 & 0.1498 & 0.0504 & 0.0329 \\
        \hline
        \wrap{Ours separate $\rho$}& 0.0452 & 0.1470 & 0.0730 & 0.1397 & 0.1622 & 0.1919 & 0.0649 & 0.0726 & 0.1241 & 0.0504 & 0.0333\\
        \hline
        \wrap{Ours No Decomposition} & 0.0436 &  0.1994 & 0.0965 & 0.1311 & 0.1654 & 0.2079 & 0.0656 & 0.0797 & 0.1226 & 0.0540 & 0.0341\\
        \hline
        \wrap{Ours No Supervision} & 0.0430 & 0.1459 & 0.0686 & 0.1346 & 0.1674 & 0.1850 & 0.0660 & 0.0762 & OOM & 0.0539 & 0.0355 \\
        \hline
        \hline
        \wrap{TensoIR} &  0.0177 & 0.2166 & 0.0724 & N/A & 0.1360 & 0.1586 & 0.0791 &0.037 & 0.045 & 0.041 & 0.039\\
        \hline
        \wrap{NMF} & 0.010 & 0.104 &  0.034 &  0.046 &  0.069 & 0.055 & 0.026 & 0.024 & 0.046 & 0.044 & N/A\\
        \hline
    \end{tabular}
    
    \label{tab:vs-nvs-lpips}
\end{table*}

\begin{table*}[t]
    \caption{MAE $\downarrow$ on geometry normal extraction (lower is better).}
    \centering
    \begin{tabular}{|c| c c c c c c c|c c c c|}
        \hline
        \wrap{Method}& Teapot& Toaster & Car & Ball & Coffee & Helmet & Materials & Lego & Hotdog & Ficus & Armadillo \\
        \hline
        \wrap{Ours} & 0.38 & 2.45 & 1.51 & 0.47 &  3.62 & 1.19 & 1.40 & 5.97 & 3.16 & 3.78 & 1.85 \\
        \hline
        \wrap{Ours separate $\rho$} & 0.38 & 2.54 & 1.50 & 0.46 &  3.55 &  1.18 & 1.37 & 5.41 & 3.37 & 3.75 & 1.83 \\
        \hline
        \wrap{Ours No Decomposition} & 0.44 & 6.58 &  1.68 & 0.36 & 3.14 & 3.17 & 1.43 & 5.45 & 3.33 & 3.94 & 1.90 \\
        \hline
        \wrap{Ours No Supervision} & 0.40 & 2.50 & 1.49 & 0.44 & 3.75 & 1.23 & 1.41 & 5.43 & OOM & 3.90 & 2.05 \\
        \hline
        \hline
        \wrap{TensoIR} & 1.05 & 8.17 & 2.90 & N/A & 4.57 & 7.45 & 3.02 & 5.980 & 4.050 & 4.420 &1.950\\
        \hline
        \wrap{NMF} & 0.752 & 4.474 & 2.598 & 1.563 & 5.352 & 1.924 & 2.868 & 8.452 & 3.546 & 4.949 & N/A\\
        \hline
    \end{tabular}
    \label{tab:vs-nvs-mae}
\end{table*}

\begin{table}[!ht]
    \caption{Tests done in with our method to compare against TensoIR in the relighting task}
    \centering
    \begin{tabular}{|l|l|l|l|l|}
    \hline
        Ours & Armadillo & Ficus & Lego & Hotdog \\ \hline
        \textbf{W/ Forest Envmap} & ~ & ~ & ~ & ~ \\
        PSNR & 34.02 & 25.57 & 26.76 & 28.93 \\ \hline
        SSIM & 0.9733 & 0.9541 & 0.9324 & 0.9249 \\ \hline
        LPIPS & 0.0475 & 0.0657 & 0.0942 & 0.1204 \\ \hline
        \textbf{W/ City Envmap} & ~ & ~ & ~ & ~ \\
        PSNR & 30.85 & 24.4 & 24.75 & 27.05 \\ \hline
        SSIM & 0.9658 & 0.9385 & 0.9151 & 0.9369 \\ \hline
        LPIPS & 0.0484 & 0.0688 & 0.0919 & 0.0935 \\ \hline
        \textbf{W/ Fireplace Envmap} & ~ & ~ & ~ & ~ \\
        PSNR & 31.9 & 22.73 & 23.62 & 29.67 \\ \hline
        SSIM & 0.9535 & 0.9394 & 0.8775 & 0.8894 \\ \hline
        LPIPS & 0.0539 & 0.078 & 0.1188 & 0.1388 \\ \hline
        \textbf{W/ Bridge Envmap} & ~ & ~ & ~ & ~ \\
        PSNR & 33.59 & 24.93 & 25.57 & 26.29 \\ \hline
        SSIM & 0.9733 & 0.9487 & 0.9232 & 0.9138 \\ \hline
        LPIPS & 0.0488 & 0.0679 & 0.0984 & 0.1237 \\ \hline
        \textbf{W/ Night Envmap} & ~ & ~ & ~ & ~ \\
        PSNR & 33.49 & 24.49 & 26.96 & 31.78 \\ \hline
        SSIM & 0.9745 & 0.9552 & 0.8506 & 0.8849 \\ \hline
        LPIPS & 0.0519 & 0.0728 & 0.0826 & 0.1015 \\ \hline
    \end{tabular}
\end{table}

\begin{table*}[!ht]
    \caption{Ablation of our method in the relighting task. Note that the ``No Supervision" Ablation failed in the hotdog scene when relighting with an OOM error. This goes to show an aspect of our method that we did not explore, the resulting size of the asset obtained. For each ablation we tested on 2 different environment maps.}
    \centering
    \begin{tabular}{|l|l|l|l|l|l|l|l|l|l|l|}
    \hline
        Relight Ablation Details & ~ & ~ & ~ & ~ & ~ & ~ & ~ & ~ & ~ & ~ \\ \hline
        Ours no supervision & Armadillo & Ficus & Lego & Hotdog & Materials & Teapot & Toaster & Coffee & Car & Helmet \\ \hline
        \textbf{W/ Snow Envmap} & ~ & ~ & ~ & ~ & ~ & ~ & ~ & ~ & ~ & ~ \\
        PSNR & 28.96 & 22.98 & 27.01 & OOM & 25.37 & 28.94 & 20.8 & 18.83 & 29.16 & 25.98 \\ \hline
        SSIM & 0.9547 & 0.9338 & 0.9239 & OOM & 0.9383 & 0.9764 & 0.8951 & 0.8807 & 0.9497 & 0.9107 \\ \hline
        LPIPS & 0.0564 & 0.0693 & 0.1011 & OOM & 0.0688 & 0.0392 & 0.1351 & 0.1512 & 0.0596 & 0.1535 \\ \hline
        \textbf{W/ Courtyard Envmap} & ~ & ~ & ~ & ~ & ~ & ~ & ~ & ~ & ~ & ~ \\
        PSNR & 30.44 & 24.45 & 25.89 & OOM & 25.46 & 25.59 & 20.85 & 19.33 & 27.97 & 25.2 \\ \hline
        SSIM & 0.9526 & 0.9431 & 0.9179 & OOM & 0.9308 & 0.9701 & 0.8823 & 0.8567 & 0.9469 & 0.898 \\ \hline
        LPIPS & 0.0556 & 0.071 & 0.0929 & OOM & 0.0631 & 0.0402 & 0.1332 & 0.1327 & 0.0586 & 0.1571 \\ \hline
        ~ & ~ & ~ & ~ & ~ & ~ & ~ & ~ & ~ & ~ & ~ \\ \hline
        Ours no decomp & Armadillo & Ficus & Lego & Hotdog & Materials & Teapot & Toaster & Coffee & Car & Helmet \\ \hline
        \textbf{W/ Snow Envmap} & ~ & ~ & ~ & ~ & ~ & ~ & ~ & ~ & ~ & ~ \\
        PSNR & 28.94 & 22.95 & 27.19 & 28.33 & 25.39 & 29.29 & 16.34 & 19.46 & 28.69 & 23.63 \\ \hline
        SSIM & 0.9562 & 0.9337 & 0.9234 & 0.8857 & 0.9391 & 0.9763 & 0.8532 & 0.8763 & 0.938 & 0.8928 \\ \hline
        LPIPS & 0.0546 & 0.0694 & 0.1038 & 0.1429 & 0.0688 & 0.0392 & 0.1834 & 0.1531 & 0.0782 & 0.1721 \\ \hline
        \textbf{W/ Courtyard Envmap} & ~ & ~ & ~ & ~ & ~ & ~ & ~ & ~ & ~ & ~ \\
        PSNR & 30.38 & 24.49 & 26.06 & 26.53 & 25.51 & 25.79 & 19.21 & 19.96 & 27.16 & 22.41 \\ \hline
        SSIM & 0.9543 & 0.9428 & 0.9178 & 0.8917 & 0.9316 & 0.9698 & 0.8488 & 0.8564 & 0.9345 & 0.8776 \\ \hline
        LPIPS & 0.0536 & 0.071 & 0.0934 & 0.1423 & 0.0623 & 0.0409 & 0.1758 & 0.1325 & 0.0785 & 0.1763 \\ \hline
        ~ & ~ & ~ & ~ & ~ & ~ & ~ & ~ & ~ & ~ & ~ \\ \hline
        Ours rho sep & Armadillo & Ficus & Lego & Hotdog & Materials & Teapot & Toaster & Coffee & Car & Helmet \\ \hline
        \textbf{W/ Snow Envmap} & ~ & ~ & ~ & ~ & ~ & ~ & ~ & ~ & ~ & ~ \\
        PSNR & 28.74 & 23.14 & 25.41 & 28.33 & 25.45 & 29.6 & 20.82 & 16.98 & 25.87 & 27.12 \\ \hline
        SSIM & 0.9585 & 0.9334 & 0.9226 & 0.8862 & 0.9397 & 0.9755 & 0.8965 & 0.8737 & 0.9385 & 0.9124 \\ \hline
        LPIPS & 0.0524 & 0.0679 & 0.0969 & 0.1429 & 0.0677 & 0.0401 & 0.1371 & 0.1563 & 0.0621 & 0.1575 \\ \hline
        \textbf{W/ Courtyard Envmap} & ~ & ~ & ~ & ~ & ~ & ~ & ~ & ~ & ~ & ~ \\
        PSNR & 30.14 & 24.6 & 24.39 & 26.63 & 25.6 & 27.78 & 20.91 & 17.53 & 27.39 & 25.82 \\ \hline
        SSIM & 0.9563 & 0.9421 & 0.9085 & 0.8919 & 0.9319 & 0.9716 & 0.8842 & 0.8525 & 0.9445 & 0.8993 \\ \hline
        LPIPS & 0.0523 & 0.0702 & 0.0962 & 0.1409 & 0.0626 & 0.0389 & 0.1377 & 0.1356 & 0.0603 & 0.1603 \\ \hline
        ~ & ~ & ~ & ~ & ~ & ~ & ~ & ~ & ~ & ~ & ~ \\ \hline
        ~ & ~ & ~ & ~ & ~ & ~ & ~ & ~ & ~ & ~ & ~ \\ \hline
        Ours & Armadillo & Ficus & Lego & Hotdog & Materials & Teapot & Toaster & Coffee & Car & Helmet \\ \hline
        \textbf{W/ Snow Envmap} & ~ & ~ & ~ & ~ & ~ & ~ & ~ & ~ & ~ & ~ \\
        PSNR & 28.85 & 22.67 & 25.14 & 28.72 & 25.34 & 29.77 & 21.3 & 16.84 & 28.53 & 27.27 \\ \hline
        SSIM & 0.9585 & 0.9329 & 0.9188 & 0.8878 & 0.9394 & 0.9803 & 0.8995 & 0.8727 & 0.949 & 0.9313 \\ \hline
        LPIPS & 0.0524 & 0.0686 & 0.1019 & 0.1423 & 0.0679 & 0.0372 & 0.1327 & 0.1561 & 0.0594 & 0.1483 \\ \hline
        \textbf{W/ Courtyard Envmap} & ~ & ~ & ~ & ~ & ~ & ~ & ~ & ~ & ~ & ~ \\
        PSNR & 30.24 & 24.02 & 23.72 & 27.16 & 25.41 & 27.91 & 21.08 & 17.41 & 27.93 & 25.91 \\ \hline
        SSIM & 0.9565 & 0.9409 & 0.9041 & 0.8959 & 0.9309 & 0.9761 & 0.8876 & 0.8533 & 0.947 & 0.9192 \\ \hline
        LPIPS & 0.0523 & 0.0709 & 0.1074 & 0.1393 & 0.0631 & 0.0356 & 0.1338 & 0.1348 & 0.0588 & 0.147 \\ \hline
    \end{tabular}
\end{table*}

%
%
%
\bibliographystyle{splncs04}
\bibliography{bib_CGI}